\documentclass[a4paper,fleqn]{cas-dc}

\usepackage[numbers]{natbib}

\begin{document}
\let\WriteBookmarks\relax
\def\floatpagepagefraction{1}
\def\textpagefraction{.001}
\shorttitle{Investigating bankruptcy prediction models in the presence of extreme class imbalance and multiple stages of economy}
\shortauthors{Islam et~al.}

\title [mode = title]{Investigating bankruptcy prediction models in the presence of extreme class imbalance and multiple stages of economy}                      

\author{Sheikh Rabiul Islam}
\cormark[1]
\author{William Eberle}
\author{Sheikh K. Ghafoor}
\author{Sid C. Bundy}
\author{Douglas A. Talbert}
\author{Ambareen Siraj}
\address{Tennessee Technological University, 1 William L Jones Dr, Cookeville, TN 38505}

\cortext[cor1]{Corresponding author}

\begin{abstract}
In the area of credit risk analytics, current Bankruptcy Prediction Models (BPMs) struggle with (a) the availability of comprehensive and real-world data sets and (b) the presence of extreme class imbalance in the data (i.e., very few samples for the minority class) that degrades the performance of the prediction model. Moreover, little research has compared the relative performance of well-known BPM's on public datasets addressing the class imbalance problem. In this work, we apply eight classes of well-known BPMs, as suggested by a review of decades of literature, on a new public dataset named Freddie Mac Single-Family Loan-Level Dataset with resampling (i.e., adding synthetic minority samples) of the minority class to tackle class imbalance. Additionally, we apply some recent AI techniques (e.g., tree-based ensemble techniques) that demonstrate potentially better results on models trained with re-sampled data. In addition, from the analysis of 19 years (1999-2017) of data, we discover that models behave differently when presented with sudden changes in the economy (e.g., a global financial crisis) resulting in abrupt fluctuations in the national default rate. In summary, this study should aid practitioners/researchers in determining the appropriate model with respect to data that contains a class imbalance and various economic stages. 
\end{abstract}




\maketitle


\section{Introduction}\label{sec:introduction}
A Bankruptcy Prediction Model (BPM) predicts bankruptcy and the financial distress of firms or lenders. In particular, the main focus of a BPM is to predict the probability that the customer will be in default or bankrupt in the near future. The high rate of bankruptcy heavily affects firm owners, partners, society, and the overall economic condition of the country at large scale (\citeauthor{alaka2018systematic}, \citeyear{alaka2018systematic}). There are several application areas for default prediction. Some of these include corporate bankruptcy, consumer mortgage default, and consumer credit card default; these application areas borrow models and domain knowledge from each other. In this work, we focus on consumer bankruptcy, and in particular, mortgage default prediction. In mortgage default prediction, a customer defaults when they are unable (or unwilling) to pay the lender for a consecutive number of periods (usually 90 days). Furthermore, the default status is followed by foreclosure, where the lender takes ownership of the home/property, and subsequently the individual can file for bankruptcy protection that allows the borrower to stop making payment to the lenders. From the algorithmic point of view, models for corporate and consumer bankruptcy prediction are similar, where the differences are primarily in the data.  

According to  \citeauthor{alaka2018systematic} (\citeyear{alaka2018systematic}), a good BPM model should possess certain characteristics, such as: it should work well with imbalanced data, the model should not overfit the data, the variable selection method should be optimal, the model should be updatable, and the result should be transparent and interpretable. Unfortunately, there is a gap in the literature when it comes to the relative performance of popular BPMs in terms of the mentioned criteria. \citeauthor{alaka2018systematic} (\citeyear{alaka2018systematic}), also mention that, due to this gap in the literature, sometimes the BPM is chosen based on just the popularity of the model or based on the professional background of the author. Another problem in consumer bankruptcy prediction literature is the use of private datasets, which is problematic for the relative comparison of performance. Some classic work on BPM includes \citeauthor{altman1968financial} (\citeyear{altman1968financial}) and  \citeauthor{ohlson1980financial} (\citeyear{ohlson1980financial}).    

A review of decades of literature by  \citeauthor{alaka2018systematic} (\citeyear{alaka2018systematic}) and  \citeauthor{bellovary2007review} (\citeyear{bellovary2007review}) indicate that there are eight popular and promising BPM tools, two of which are Statistical: (1) Multiple Discriminant Analysis (MDA) and (2) Logistic Regression (LR); and the remaining six are Artificial  Intelligence (AI) based tools:  (1) Artificial Neural Network (ANN), (2) Support Vector Machine (SVM), (3) Rough Sets (RS), (4) Case-Based Reasoning (CBR), (5) Decision Tree (DT), and (6) Genetic Algorithm (GA). In this work, we experiment with these aforementioned algorithms, as well as a few recent popular AI techniques which are uncommon as BPM models, such as: Random Forest (RF), Extremely Randomized Trees or Extra Trees (ET), Adaptive Boosting or AdaBoost (AB), and Gradient Boosting (GB). In addition, we add a probabilistic model, Naive Bayes (NB), to our experiments as it assumes conditional independence among features. For the rest of the paper, we will use the mentioned short form of the algorithms frequently. An intuitive description for each of these models/algorithms is available in section \ref{subsec:prediction_models}. We will use the terms \textit{algorithm} and \textit{model} interchangeably. 

To mitigate some of the mentioned gaps in the literature, we use a public and comparatively new (first published in 2013 and updated continuously after some interval) dataset called the "Freddie Mac Single-Family Loan-Level" Dataset (or FMSFLL Data for short) for a comprehensive analysis of all of the models. To increase transparency and improve BPMs, Freddie Mac, a government-sponsored enterprise, is making available loan-level credit loan origination and performance data on fixed-rate mortgages that the company purchased or guaranteed from 1999 to 2017. 

We will implement and run all BPMs using similar configurations on the same data, over different economic cycles over 19 years of time (1999 to 2017). We divide these 19 years into three different stages of the economy and show the sensitivity (i.e., recall) of the models as the economy and class distribution changes. In this  single comprehensive work on all well-known BPMs, we attempt to fill in the literature gap by performing the following: (a) we oversample the minority class using a popular technique called SMOTE (\citeauthor{chawla2002smote}, \citeyear{chawla2002smote}) as class imbalance degrades performance of the BPM and the Freddie Mac dataset is extremely imbalanced; (b) we select an optimal set of features using different methods (e.g., Random Forest, Correlation analysis, Genetic Algorithm) and use those features for all models; (c) we use the same dataset, same training vs test set ratio for a comprehensive set of models; and (d) to test the model, we use the same holdout set that is never shown to any of the models during the training phase. 

We find that, in terms of the important performance metric for imbalanced data (e.g., recall, ROC-AUC), almost all models tends to show better performance when the minority class of the training dataset is oversampled. We also find that the level of imbalance  has performance issues in the different economic stages. We expect that this study will help practitioners with an understanding of the pros/cons of different models with respect to class imbalance and economic stages for developing/tuning their BPM model. 

We start with a background of related work (Section \ref{sec:background}) followed by intuitive descriptions of all methods and algorithms including an overview of the dataset (Section \ref{sec:methods}) used in this work. In Section \ref{sec:experiments}, we describe our experiments, followed by Section \ref{sec:results_and_descussions}, which contains results and discussions. We conclude with limitations and future work in Section \ref{sec:conclusions}.

\section{Background}\label{sec:background}
Since the 1960s, research in the area of default prediction and bankruptcy prediction models have focused on corporate bankruptcy prediction. More recently, default prediction models for consumer credit were derived from advances in corporate bankruptcy prediction models. However, the models are similar, with the primary differences being in explanatory variables and the data. We divide the background work into two main subsections based on the relevance and nature of the work: corporate credit risk models and  consumer credit risk models. Finally, we focus on class imbalance irrespective of consumer/corporate credit risk models.

\subsection{Corporate Credit Risk}\label{subsec:corporate_credit_risk}
In 1968, Altman's seminal Z-Score (\citeauthor{altman1968financial}) used Multiple Discriminant Analysis (MDA) to predict bankruptcy using key ratios based entirely from publicly available financial information. This model is based on Beaver's (1966) (\citeauthor{beaver1966financial}) recommendation in his univariate work. In 1980,  \citeauthor{ohlson1980financial} updated the statistical model for an MDA model by providing evidence that a logistic-regression-based BPM outperforms Altman's Z-Score for corporate bankruptcy prediction. His model includes seven ratios that come not only from financial statement information but also from economic conditions (e.g., total asset vs price index for Gross National Product (GNP)). Based upon this work, outside information and indicator variables began to be incorporated into BPMs. Later, in 1986  \citeauthor{lane1986application}, used Linear Discriminant Analysis (LDA) and Quadratic Discriminant Analysis (QDA) as a BPM. In 1995, \citeauthor{boritz1995effectiveness} came up with a neural network based BPM that uses 14 variables/factors though, in 1990, Bell et al., \cite{bell1990neural} use Neural Networks for a BPM, outperforming the \textit{Logit} model. According to \citeauthor{bellovary2007review} (\citeyear{bellovary2007review}), the number of factors considered for BPM models ranges from one to 57. Later, in 1996, \citeauthor{bryant1997case} came up with a Case-Based Reasoning BPM, and in 1999 \citeauthor{dimitras1999business} came up with a Rough Set Theory based BPM that uses 12 factors for corporate bankruptcy prediction.

\subsection{Consumer Credit Risk}\label{subsec:consumer_credit_risk}
In terms of consumer credit risk, we particularly focus on mortgage bankruptcy prediction. There has been a substantial amount of research on mortgage delinquency. In 1969,  \citeauthor{von1969default} formulated the influence of variables such as income, loan age, and loan-to-value ratio on the default rate. Later,  \citeauthor{schwartz1993mortgage} introduced the use of macro-economic variables along with loan-level variables. 

\citeauthor{anderson2014building} (\citeyear{anderson2014building}) show the inclusion effect of different macro-economic (e.g., unemployment rate, housing prices) variables and credit underwriting environments across different credit origination vintages.  \citeauthor{goodman2014look} (\citeyear{goodman2014look}) used the Freddie Mac dataset and found that lower FICO scores and a higher Loan To Value (LTV) increases the likelihood of default. 

\citeauthor{sousa2015links} (\citeyear{sousa2015links}), also uses the same Freddie Mac dataset, and investigates two mechanisms of memory for credit risk assessment: short-term memory (STM) and long-term memory (LTM). They found that STM consistently outperforms LTM due to its quick adaption to changes. In short, newer information helps improve accuracy compared to older information.   

In addition, \citeauthor{sousa2016new} (\citeyear{sousa2016new}) also investigate the dynamics and performance of over 16.7 million fully amortized loans from the Freddie Mac dataset. Their findings reveal that research in credit risk assessment lacks validation that is representative of a real-world environment, and the dataset used in the experimental design is not representative of each economic cycle/phase, which leads to a lack of generalization ability for a significant portion of empirical studies. In addition, traditional static models are one shot models with fixed memory, not capable of dealing with highly evolving data due to concept drift (i.e., regulatory movements, interest rates fluctuations). In this work, from our analysis, we also found that the default rate changes with a different phase of the economy which leads to changes in class distribution in the data. In both works by Sousa et al. (\citeauthor{sousa2015links}, \citeauthor{sousa2016new}), the main focus was not only predicting default, but also on concept drift and determining the appropriate amount of data needed for a particular model for optimal prediction.  

A comprehensive work for default prediction using deep learning was carried out by \citeauthor{sirignano2016deep} (\citeyear{sirignano2016deep}) on the CoreLogic dataset of 120 million prime and subprime mortgages. Their model outperforms logistic regression by 10\% on the ROC curve. They also found that the inclusion of macroeconomic variables improves predictive power. 

\subsection{Class Imbalance}\label{subsec:class_imbalance}
\citeauthor{garcia2012improving} (\citeyear{garcia2012improving}) study a number of re-sampling techniques applied over five public credit datasets, using four classification algorithms: Multi-Layer Perceptron (MLP), Support Vector Machines (SVM), 1-Nearest Neighbor (NN), and Radial Basis Function (RBF).  They found that irrespective of the classification algorithm, the re-sampled dataset shows better performance. However, unlike the work presented in this paper, which covers a broad class of twelve different algorithms and highly imbalanced data (minority class < .2\%), the minimum amount of imbalance handled in any of their datasets is 5.26\%. Moreover, our analysis covers the effect of class imbalance in different economic phases. In addition, they present the results of their work using AUC, without any discussion of recall and precision\textemdash{}something we deem to be crucial (see Section \ref{subsec:evaluation_criteria}) to measure the performance for imbalanced classes. 

Similarly, \citeauthor{brown2012experimental} (\citeyear{brown2012experimental}), provide a comparative study of different credit scoring techniques. However, they under-sample the majority class without using any re-sampling techniques that create artificial/synthetic samples. Under-sampling suffers from the problem of discarding some instances which are valuable for the model leading to over-fitting or under-fitting situations. Their selection of an algorithm for an experiment doesn't include any of the ensemble techniques (e.g., RF, ET, Boosting) that tend to show better performance in many problems. Similar to \citeauthor{garcia2012improving} (\citeyear{garcia2012improving}), they use AUC as the performance metric, instead of using recall to better evaluate customer defaults. 

The work of \citeauthor{chen2016financial} (\citeyear{chen2016financial}) is a review of current credit risk assessment research. Similarly,  \citeauthor{sun2009classification} (\citeyear{sun2009classification}) provides a review of classification of imbalanced data in general. They address imbalanced data from multiple directions: re-sampling the data space, using boosting approaches, cost-sensitive learning, and adapting existing algorithms by introducing learning biases towards the target class (e.g., fraudulent class). However, these review papers do not show any experimental results or comparison of different models to see how sensitive they are to the class imbalance problem. 

\citeauthor{sayli2010comparative} (\citeyear{sayli2010comparative}) provide an analysis of five classes of classification techniques on different credit datasets: DT, LR, SVM, ANN, and RF. They also over-sample (duplicating) the minority class without using any re-sampling techniques. In addition, their datasets are not highly imbalanced, as the minimum percentage for the minority class is 20\%. Besides, accuracy is the only performance metric used to evaluate the performance of the classifier, which is not a good metric to measure the performance of the imbalanced class. 

The following is a comprehensive study that fills in some of the gaps found in previous work, including the use of proper metrics for measuring the performance for imbalanced classes, the use of highly imbalanced datasets, considering different economic stages, and the use of a re-sampling technique that does not just create a duplicate sample, rather artificial/synthetic samples for the minority class that help to avoid the over-generalization or over-fitting problem.

\section{Methods}\label{sec:methods}
We begin with an intuitive and brief discussion of all thirteen prediction models, provide a description of data and data preparation strategies, then follow with the procedure of generating synthetic data (see Section \ref{subsubsection:smote}) for minority class oversampling using SMOTE (\citeauthor{chawla2002smote}, \citeyear{chawla2002smote}).

\subsection{Prediction models}\label{subsec:prediction_models}
\subsubsection{Artificial Neural Network (ANN)}\label{subsubsection:ann}
An Artificial Neural Network is a non-linear model, capable of mimicking human brain functions. It consists of an input layer, multiple hidden layers, and the output layer. Each layer consists of multiple neurons that help to learn complex patterns, where each layer following the previous layer learns more abstract concepts before it merges into the output layer. ANN was first used in 1994 by \citeauthor{wilson1994bankruptcy} for bankruptcy prediction. Usually, ANN is a data-hungry model and doesn't work well with smaller datasets. It also comes with comparatively more training time and hyper-parameter tuning requirements. In this work, we use GridSearchCV, part of the scikit-learn library, to find optimal hyper-parameters for the model. It is still limited by the given range of parameters and is very time consuming as it tries all possible combinations of the parameter in the given range. In terms of accuracy, given enough data, ANN performs best for many of the problems due to its capability of learning any non-linear function. 

\subsubsection{Multiple Discriminant Analysis (MDA)}\label{subsubsection:mda}
Multiple Discriminant analysis is a multivariate technique that reduces multiple measurements into a single composite score to classify two or more groups. It projects high-dimensional data onto a line in one-dimensional space and then classifies the groups (\citeauthor{fisher1936use}, \citeyear{fisher1936use}).  

Furthermore, Fisher's Discriminant Analysis (\citeauthor{fisher1936use}, \citeyear{fisher1936use}), is similar to multiple regression; it comes with only two class classification though. Linear Discriminant Analysis (LDA) is an extension of Fisher's Discriminant Analysis for the classification task that uses matrix decomposition (e.g., eigen decomposition) for the classification problem. A more advanced multiple discriminant analysis technique is Quadratic Discriminant Analysis (QDA), which is a generalization of LDA. Both use Gaussian assumption but LDA assumes equal variance-covariance matrices of the input variables for classes, leading to a linear decision boundary whereas QDA allows different covariance matrices for different classes leading to a quadratic decision boundary. If we are using LDA for a problem where their covariance matrices differ noticeably then the majority of the data will be classified as the class with higher variability. QDA is a good candidate in this case as it allows heterogeneity of covariance matrices for classes. We have used scikit-learn Quadratic Discriminant Analysis in our experiment that gives a quadratic decision boundary using class conditional densities of data and uses Bayes rule \cite{sklean_discriminant}. 

\subsubsection{Rough Sets (RS)}\label{subsubsection:rs}
Rough Set Theory (RST), proposed by \citeauthor{pawlak1982rough} (\citeyear{pawlak1982rough}), is the first non-statistical approach in data analysis concerned with classification and analysis of imprecise, uncertain or incomplete knowledge and information. RST obviates the requirement for additional information (i.e., probability distribution) or the value of possibility as required in fuzzy set theory (\citeauthor{rissino2009rough}, \citeyear{rissino2009rough}). It also helps to approximate the description of a concept. For example, the rough set can be used to approximate a description of the fuzzy concept where the rule patterns can be represented in a more compressed way. The rough k-means algorithm is based on rough set theory, where a cluster is represented by a rough set of lower approximation and upper approximation (\citeauthor{kumar2011comparative}, \citeyear{kumar2011comparative}). Here are the basic properties of the algorithm (\citeauthor{rissino2009rough}, \citeyear{rissino2009rough}):
\begin{itemize}
    \item Lower approximation space consists of objects that are \textit{definitely} part of an interest subset. An object can be only in one lower approximation of a cluster. It can also be part of the upper approximation of the same cluster. 
    \item Upper approximation space consists of objects that are  \textit{possibly} part of an interest subset.
    \item An object that doesn't belong to any lower approximation is a member of at least two upper approximation. 
    \item Every subset defined through lower and upper approximation space is called Rough Set. 
\end{itemize}

The following is how the algorithm works (\citeauthor{kumar2011comparative}, \citeyear{kumar2011comparative}):
\begin{enumerate}
    \item Select some initial clusters (k clusters).
    \item Assign each object to lower approximation space or upper approximation space of cluster/clusters respectively as follows:
    \begin{itemize}
        \item If the size of the gap between two clusters center from an object is less than the threshold then the object can't belong to any of respective lower bound; it will be part of upper bound of both clusters. 
        \item Otherwise, the object will be part of the lower approximate of the closest cluster.
    \end{itemize}
    \item For each cluster, re-compute cluster center according to the weighted combination of objects in its lower approximation and upper approximation. 
    \item Stop in case of convergence (i.e., cluster center doesn't move anymore).
\end{enumerate}

\citeauthor{kumar2011comparative} (\citeyear{kumar2011comparative}) applied rough k-means on a cancerous gene expression dataset to detect Leukemia and found that rough k-means algorithm deals with uncertainty in a better way than k-means and other algorithms used in their experiment.

\subsubsection{Case Based Reasoning (CBR)}\label{subsubsection:cbr}
During the 70s and 80s, rule-based expert systems (RBES) were very popular despite some of the limitations such as (a) building a knowledge base requires extensive domain knowledge and is time consuming, (b) inability to deal with problems that doesn't explicitly align with the utilized rule base, (c) frequent requirements of  programmer intervention to cope with new problems or requirements. In the following decade, a new methodology, Case Based Reasoning (CBR) solves above-mentioned problems, a CBR solves the problem by using or adapting solutions that were used to solve old problems (\citeauthor{riesbeck2013inside}, \citeyear{riesbeck2013inside}). The basic principles of CBR are as follows:
\begin{enumerate}
    \item Retrieve similar cases that match the description of the new problem.
    \item Reuse a solution that was suggested for one of the similar cases.
    \item Revise or adapt that solution for a better fit for future problems.
    \item Retain the solution after validation. 
\end{enumerate}
\citeauthor{watson1999case} (\citeyear{watson1999case}) argued that CBR is not a specific technology (e.g., Neural Networks), rather it is a methodology that has various use: nearest neighbor, induction, fuzzy logic and SQL. The nearest neighbor technique is the most widely used CBR technique where similarity of a target case is determined against a case in the library for each of its attributes. CBR methodology includes induction algorithms, such as decision trees (e.g., ID3) that identify patterns among cases and partition the cases into clusters based on similarity.  CBR also includes fuzzy logic based technique that represents notions of similarity, for example, a feature can be represented as excellent, good, fair, and poor to refer the differences since good is closer or more similar to excellent than it is to poor.

Furthermore, CBR uses a fuzzy preference function to measure the similarity with an attribute of a new case with an existing case's corresponding attribute, it allows measuring a smooth change in attribute values. In addition, CBR can also be implemented using database technology as it is an efficient means of storing and retrieving a large volume of data. The problem is expressed as a well-formed query to find a similar case from the database. But using only straight forward SQL, it retrieves cases with exact matches only. To overcome these issues, augmenting a database with explicit domain knowledge about the problem can enable SQL queries to find similarities. Historically CBR has shown low accuracy in BPM due to the failure of taking feature importance into account (\citeauthor{chuang2013application}, \citeyear{chuang2013application}). Furthermore, it is unable to handle the non-linear problem. In a few research studies, the hybrid solution of CBR performed better, though overall it is a less accurate model in general (\citeauthor{alaka2018systematic}, \citeyear{alaka2018systematic}). 

We have excluded CBR from our experiment, due to its low accuracy and varieties of implementation. In fact, as mentioned earlier, it is a methodology rather than a standalone technique. Moreover, it incorporates some of the technique (e.g., DT, nearest neighbor, fuzzy logic) and none of those are in our top eight standalone algorithms for BPM.	

\subsubsection{Logistic Regression (LR)}\label{subsubsection:lr}
Logistic Regression (LR) is a model that is capable of predicting a categorical (e.g., binary) decision from a linear combination of predictor variables (continuous or categorical). For example, predicting whether a customer will fall into the default or non-default category given that customer's current credit score, income, and the value of the house. While LR can generate a simple probabilistic model for classification, it doesn't work well with the non-linear problem. LR was developed by  \citeauthor{cox1958regression} (\citeyear{cox1958regression}), since then it has been extensively used in credit risk analytics (\cite{hooman2016statistical}, \citeauthor{hooman2016statistical}).

\subsubsection{Decision Tree (DT)}\label{subsubsection:dt}
DT became a popular machine learning tool when \citeauthor{quinlan1986induction} (\citeyear{quinlan1986induction}) developed Iterative Dichotomiser 3 (ID3) that uses entropy to measure the discriminative power of a feature (\citeauthor{alaka2018systematic}, \citeyear{alaka2018systematic}). The feature with most discriminative power is placed at the top of the tree and iteratively other features are placed accordingly towards the bottom of the tree in a top-down fashion. Some of the advantages of DT include: easy to understand and interpret, not data hungry, and can handle both numerical and categorical data. However, a small variation in the data might result in considerably different trees. Tree-based ensemble approaches help to tackle this issue though. In addition, Non-pruned DT can overfit data easily. 

\subsubsection{Random Forest (RF)}\label{subsubsection:rf}
Random Forest is a tree-based ensemble technique developed by  \citeauthor{breiman2001random} (\citeyear{breiman2001random}) for the supervised classification task. In RF, many trees are generated from  bootstrapped subsamples (i.e., random sample drawn with replacement) of training data. In each tree, the splitting attribute is chosen from a smaller random subset of attributes for that tree (i.e., the chosen split attribute is best among that random subset); this randomness helps to make the trees less correlated, which is good because correlated trees make the same kinds of prediction error and overfit the model as well. By this technique, a forest of trees is built, and the output from all the trees are averaged to make the final prediction. This averaging helps to reduce the variance from the model. Furthermore, RF can works with a parallel computing environment as trees can be grown independently. So far, RF has been used in different credit scoring and customer attrition applications (\citeauthor{fitzpatrick2016empirical}, \citeyear{fitzpatrick2016empirical}). 

\subsubsection{Extra Trees (ET)}\label{subsubsection:et}
Extremely Randomized Trees or Extra Trees (ET) is also a tree-based ensemble technique like RF and shares a similar concept with Random Forest (RF). The only differences are in the process of selecting the splitting attribute and in determining the threshold (cutoff) value; both are chosen in extremely random fashion (\citeauthor{islam2018mining}, \citeyear{islam2018mining}). As in RF, a random subset of features is taken into consideration for the split selection but instead of choosing the most discriminative cut off threshold, here in ET, initially the cut off thresholds are set to random values. By this way, at the end, the best of these randomly chosen value is set as the threshold for the splitting rule (\citeauthor{ensemble_methods}, \citeyear{ensemble_methods}). As a result of multiple trees, the variance reduces a little bit compared to Decision Trees, however, as a subset of the whole feature set is chosen for each tree, that comes with little increase of bias. The ET which was proposed by \citeauthor{geurts2006extremely} (\citeyear{geurts2006extremely}), has continued its success by achieving the state of the art performance in some anomaly/intrusion detection research (\citeauthor{islam2018efficient}, \citeyear{islam2018efficient}; \citeauthor{islam2018credit}, \citeyear{islam2018credit}). 

\subsubsection{AdaBoost (AB)}\label{subsubsection:ab}
AdaBoost is also known as Adaptive Boosting (AB) algorithm, is a tree-based ensemble technique where each tree is run sequentially, and each subsequent attempt tries to fix the errors made by its predecessors. It gradually makes weak learners stronger by boosting the weight of misclassified instances so that in the next model pays more attention to those. AB was proposed by \citeauthor{freund1997decision} (\citeyear{freund1997decision}), is relatively new compared to other algorithms, and has already shown promising performance in some applications. 

\subsubsection{Gradient Boosting (GB)}\label{subsubsection:gb}
\citeauthor{friedman2001greedy} (\citeyear{friedman2001greedy}), generalized Adaboost to Gradient Boosting algorithm that allows for a variety of loss function. Here the shortcoming of weak learners is identified using the gradient, while in AdaBoost it is done through highly weighted data points.  Gradient Boosting (GB) is a classifier/regression model in the form of an ensemble of weak prediction models, usually, Decision Trees are fitted with data initially as weak learners. It also works sequentially like the AdaBoost algorithm. In each subsequent model it tries to minimize the loss function (i.e., Mean Squared Error) by focusing on instances that were hard to get right in previous steps. 

\subsubsection{Support Vector Machine (SVM)}\label{subsubsection:svm}
The Support Vector Machine (SVM) algorithm was first introduced by \citeauthor*{boser2003training} (\citeyear{boser2003training}) and is used for supervised classification tasks. The model learns an optimal hyperplane that separates instances of different classes using a highly non-linear mapping of input vectors in high dimensional feature space (\citeauthor{hooman2016statistical}, \citeyear{hooman2016statistical}).  SVM is listed as one of the top non-linear algorithms for bankruptcy prediction in different literature surveys (\citeauthor{alaka2018systematic}, \citeyear{alaka2018systematic}; \citeauthor{bellovary2007review}, \citeyear{bellovary2007review}). When the number of samples is too high (i.e., millions) then it is very costly in terms of computation time. In that case, a data-hungry and non-linear algorithm like ANN can be a better choice.

\subsubsection{Genetic Algorithm (GA)}\label{subsubsection:ga}
The Genetic Algorithm technique was first proposed by \citeauthor{holland1975adaptation} in 1960, is inspired by Darwin's theory of Natural Selection (\citeauthor{edelman1987neural}, \citeyear{edelman1987neural}), and is mostly used for optimization problems where there is incomplete or imperfect information or limited computing capacity. The three main principles of natural selection are: there must be a way for children to receive traits of parents,  there must be a way to introduce variations of traits in the population, and there must be a way to select some individual as a parent based on likelihood of survival (i.e., fitness score). So, a Genetic Algorithm can be implemented using the following steps:
\begin{itemize}
    \item creating the initial population, defining the fitness function that calculate fitness score of each individual;
    \item selecting parents based on the fitness score, and passing genes of those parents to the next generation;
    \item introducing variations by crossing over (e.g., permutations, combinations) of different survived populations in the new generation;
    \item adding little bit of mutation/alteration to avoid premature convergence\textemdash{}a few of the new population (e.g., .5\%) still remain the same as it was in the previous generation due to the effect of crossover (e.g., permutation or combination); and
    \item finally when no more new unique population is generated which is significantly different from previously seen population or the algorithm has reached to a predefined number of iteration then the algorithm can stop.
\end{itemize}
Genetic Algorithms has been used for bankruptcy prediction from long before. In this work, we combined Genetic Algorithm with Random Forests to fit it with a supervised problem\textemdash{}genetic algorithm provides key features from the dataset and finally we run the RF using the selected features to make it a supervised model. 

\subsubsection{Naive Bayes (NB)}\label{subsubsection:nb}
The Naive Bayes algorithm is based on Bayes Theorem which was formulated in the seventeenth century. It is a supervised, simple, and comparatively fast algorithm based on statistics. In real-world problems, it is unusual that all features are independent. However, naive Bayes assumes conditional independence among features and surprising works well in most cases. This assumption of Naive Bayes helps to avoid lots of computations (e.g., computing the conditional probability for each feature with others) and makes it a faster algorithm. 

\subsection{Data}\label{subsec:data}
To increase transparency, support risk-sharing initiatives, and build more accurate credit performance models, Freddie Mac, a government-sponsored enterprise, is making available loan-level credit performance data on fixed-rate mortgages that the company purchased or guaranteed from 1999 to 2017. The dataset contains 26.6 million mortgages over 19 years of time, and the use of the dataset is free for non-commercial, academic, and research purposes (\citeauthor{single_family}, \citeyear{single_family}). The dataset is organized into 19 different folders for 19 different loan origination vintages. For each year, there are two kinds of files in the dataset (a) a loan origination file which contains the customer's detailed information (anonymized) for the loans that originated in that vintage; and (b) a performance file which contains monthly payments and other variables, updated month to month from the origination vintage to the latest known status of the loan. So, for a particular customer, for a vintage, there is one record containing the loan origination information and multiple records (on average 45) for loan performance information. This provides the models with a one-to-many relationship amongst the data, allowing for a concatenation between the loan origination records and each performance record. Due to the huge size of the dataset, most of the previous work on this dataset used the minimal representative version of the dataset that contains 50,000 loans from each vintage year (\citeauthor{anderson2014building}, \citeyear{anderson2014building}). Table \ref{tab:dataset} shows all 50 features of the dataset.  
\begin{table*}
\caption{Freddie Mac dataset: showing both loan origination and loan performance features.}
\label{tab:dataset}
\centering
\begin{tabular}{lllll}
\toprule
SL\# & \multicolumn{2}{l}{Loan origination features}     & SL\# & Loan performance features        \\
\midrule
1    & \multicolumn{2}{l}{creditScore}                   & 1    & loanSequenceNumber               \\
2    & \multicolumn{2}{l}{firstPaymentDate}              & 2    & monthlyReportingPeriod           \\
3    & \multicolumn{2}{l}{firstTimeHomeBuyerFlag}        & 3    & currentActualUPB                 \\
4    & \multicolumn{2}{l}{maturityDate}                  & 4    & currentLoanDelinquencyStatus     \\
5    & \multicolumn{2}{l}{metropolitanDivisionOrMSA}     & 5    & loanAge                          \\
6    & \multicolumn{2}{l}{mortgageInsurancePercentage}   & 6    & remainingMonthToLegalMaturity    \\
7    & \multicolumn{2}{l}{numberOfUnits}                 & 7    & repurchaseFlag                   \\
8    & \multicolumn{2}{l}{occupancyStatus}               & 8    & modificationFlag                 \\
9    & \multicolumn{2}{l}{originalCombinedLoanToValue}   & 9    & zeroBalanceCode                  \\
10   & \multicolumn{2}{l}{originalDebtToIncomeRatio}     & 10   & zeroBalanceEffectiveDate         \\
11   & \multicolumn{2}{l}{originalUPB}                   & 11   & currentInterestRate              \\
12   & \multicolumn{2}{l}{originalLoanToValue}           & 12   & currentDeferredUPB               \\
13   & \multicolumn{2}{l}{originalInterestRate}          & 13   & dueDateOfLastPaidInstallment     \\
14   & channel &                                         & 14   & miRecoveries                     \\
15   & \multicolumn{2}{l}{prepaymentPenaltyMortgageFlag} & 15   & netSalesProceeds                 \\
16   & \multicolumn{2}{l}{productType}                   & 16   & nonMiRecoveries                  \\
17   & \multicolumn{2}{l}{propertyState}                 & 17   & expenses                         \\
18   & \multicolumn{2}{l}{propertyType}                  & 18   & legalCosts                       \\
19   & \multicolumn{2}{l}{postalCode}                    & 19   & maintenanceAndPreservationCosts  \\
20   & \multicolumn{2}{l}{loanSequenceNumber}            & 20   & taxesAndInsurance                \\
21   & \multicolumn{2}{l}{loanPurpose}                   & 21   & miscellaneousExpenses            \\
22   & \multicolumn{2}{l}{originalLoanTerm}              & 22   & actualLossCalculation            \\
23   & \multicolumn{2}{l}{numberOfBorrowers}             & 23   & modificationCost                 \\
24   & \multicolumn{2}{l}{sellerName}                    &      &                                  \\
25   & \multicolumn{2}{l}{servicerName}                  &      &                                  \\
26   & \multicolumn{2}{l}{superConformingFlag}           &      &                                  \\
27   & \multicolumn{2}{l}{preHarpLoanSequenceNumber}     &      &      \\
\bottomrule
\end{tabular}
\end{table*}

\subsection{Data Preparation}\label{subsec:data_preparation}
We cleaned, labeled, and sampled the dataset before feeding it to the model.

\subsubsection{Data cleaning}\label{subsubsection:data_cleaning}
For samples whose values for the key features such as creditScore,  originalLoanToValue, originalDebtToIncomeRatio, and originalInterestRate are missing, we simply remove those records. These features are very important, and the missing value might be the result of reporting errors or incomplete information provided by the borrower during the loan origination time. \citeauthor{sirignano2016deep} (\citeyear{sirignano2016deep}), use a different dataset (e.g., Corelogic) which includes data and features from the FMSFLL dataset and apply the same technique for missing data in important fields. For other features with missing data, for nominal values we replace the empty fields with ''Not Available'' (a separate category), and for numeric data, we replace all empty/null fields with zero, to avoid errors from ML models. Empty/Null fields are problematic for a few ML models.  

\subsubsection{Data labeling}\label{subsubsection:data_labeling}
The dataset is not directly labeled with whether accounts are default or non-default. We follow the same data labeling technique used by \citeauthor{bhattacharya2019bayesian} (\citeyear{bhattacharya2019bayesian}) for this dataset, where an account is treated as default (we name the target variable as defaulted) if the feature zeroBalanceCode = 03, 06, or 09. The feature zeroBalanceCode tells the reason for which the balance is zero. Different possible values for feature zeroBalanceCode are:
\begin{itemize}
    \item  01 = Prepaid or Matured (Voluntary Payoff)
    \item 03 = Foreclosure Alternative Group (Short Sale, Third Party Sale, Charge Off or Note Sale)
    \item 06 = Repurchase prior to Property Disposition
    \item 09 = REO Disposition\textemdash{}Real Estate Owned\textemdash{}a foreclosed mortgage that does not sell at auction, and the servicer or the lender assumes the ownership. 
    \item Double blank space or empty = Not Applicable.
\end{itemize}

\subsubsection{Data sampling}\label{subsubsection:data_sampling}
In the minimal version of the dataset, each loan origination vintage has 50,000 customers' loan origination information and on average 2,250,000 records for loan performance information (i.e., on average 45 records per customer). This minimized version of the dataset has a similar class distribution as the original dataset with a maximum difference of .4\%, which is very minimal. This is still a large number of samples (e.g., 2,250,000 per year) in terms of the computational requirement for most of the algorithms under consideration for the experiments without using any high-performance computing machines. So, we take a stratified sample of 2,000 customers from each loan vintage year out of the 50,000 customers, giving us approximately 2,000 * 45 = 90,000 records from each of 19 loan origination vintage years. We did the stratified sampling to maintain the same class ratio as in the dataset before sampling. 
\subsubsection{Oversampling using SMOTE}\label{subsubsection:smote}
It is obvious that our dataset has the class imbalance problem (e.g., in year to year, the minority class ranges from 0.01\% to .18\%, roughly <. 2\%), as one class has many more examples compared to the other classes (see section \ref{subsec:different_loan_underwriting}). After merging with performance data, our dataset becomes extremely imbalanced. In fact, almost all credit default estimation datasets are imbalanced. This class imbalance problem impairs the predictive capability of classification algorithms (\citeauthor{last2017oversampling}, \citeyear{last2017oversampling}). In the imbalanced data, the classification algorithms aim to maximize the classification accuracy which is biased toward the majority class. In other words, a classifier can achieve very high accuracy without predicting a single minority class correctly. 

\begin{figure}
  \centering
  \includegraphics[width=\linewidth]{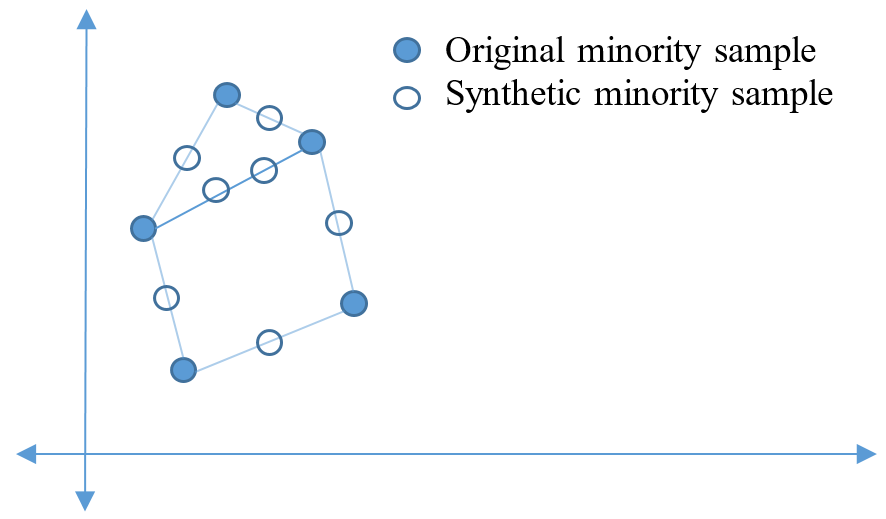}
  \caption{Oversampling using SMOTE.}
  \label{fig:smote}
\end{figure}

To overcome this problem, we use the well-known oversampling technique SMOTE (\citeauthor{chawla2002smote}, \citeyear{chawla2002smote}) to oversample the minority class. It creates synthetic samples rather than just oversampling with replacement. The minority class is oversampled by creating new examples along the line segments joining any or all of k-nearest minority samples, where k is chosen based on the percentage of oversampling required (i.e., hyperparameter to the algorithm) (\citeauthor{chawla2002smote}, \citeyear{chawla2002smote}). In Figure \ref{fig:smote}, the filled points are some minority samples. From each minority sample, SMOTE identifies its nearest neighbor, measures the difference between feature vectors, multiplies the difference with a random number between 0 and 1, and then places the new point (e.g., empty circles) on the line segment connecting two other minority samples. This process is repeated until the number of oversamples requested is met. 
We resort to a sampling technique that creates artificial/synthetic samples because sampling techniques that under-sample the data by reducing the majority class might introduce bias as discarded samples might have some important information that is no longer available. On the other hand, random oversampling that creates duplicate minority samples is prone to overfitting due to the matter that the classifier is tightly fitted to that training data, therefore it loses the generalization ability for the test data.

\section{Experiments}\label{sec:experiments}
We start with describing the experiment setup procedures, computing resources, feature selection process, oversampling procedure, and characteristics of the holdout set. Then, we describe how we investigate the effect of class imbalance (model wise) during different economic stages with different loan underwriting standards and fluctuating default rates. Finally, we discuss the evaluation of appropriate criteria for the imbalanced dataset used in this work. 

\subsection{Experimental setup}\label{subsec:experimental_setup}
Algorithms are implemented using Python, various Python packages (e.g., sci-kit learn), open-source libraries, some customized versions (e.g., rough sets, GA), and original code (e.g., ANN using Keras). We excluded CBR from our experiment as it can have several implementations. We run all experiments on a GPU-enabled desktop with 12GB RAM and a core i7 processor, running the Ubuntu operating system.  

Using the above settings, we trained twelve models using the original dataset, and tested against the holdout dataset (see Section \ref{subsubsection:holdout_set}). We also trained the same twelve models using the resampled data (using SMOTE) and tested with the same holdout dataset. Our target is to predict accounts that have a higher chance of default at least one month before to anytime in the future. The comparison of the results is in Section \ref{sec:results_and_descussions}. Before running the BPC, we reduce the feature set using different feature selection techniques and oversampled the minority class of the training dataset. We make the source code available to the research community to replicate the experiments at \cite{project_code}.

\subsubsection{Feature selection}\label{subsubsection:feature_selection}
In total, we have 50 features in the dataset from the combined origination and performance files. Too many features usually overfit the model and increase the processing complexity. So, we discard unnecessary features. First, we remove all date fields as we are not doing any time series analysis and retaining those might be deemed as noise by the model or cause overfitting. Excluded fields include the following: firstPaymentDate, maturityDate, monthlyReportingPeriod, and dueDateOfLastPaidInstallment. Then, from the remaining features, we select the primary features using two types of approaches: (1) a filter method which is independent of any machine learning algorithms, features are selected on the basis of scores from different statistical test (e.g., correlation analysis); and (2) two wrapper methods (RF and GA) that discover an optimal subset of features from all features by training a model and observing the deviations of performance for different combinations of features (e.g. using backward elimination of features) . The Genetic Algorithm (GA) selects important features based on the survived features and their fitness score in the final generation. 

The reason behind applying multiple wrapper methods for feature selection is that GA and RF tend to be stochastic, giving a slightly different ranking of features in each run. Our target is to make a list of unimportant features that can be deleted. So we crosscheck the features identified as unimportant (all features minus important features) by RF and GA, with the results of the correlation analysis, to check whether we are discarding features that have a non-negligible amount of correlation (e.g., >=.1) with our target dependent variable. Finally, features that were deemed unimportant by all three approaches were discarded. Table \ref{tab:deleted_features} is the list of features that were deleted. It also shows the associated feature importance and correlation coefficient. 

\begin{table}
\caption{Deleted features: showing features importance by RF and correlation with default for deleted features.}
\label{tab:deleted_features}
\centering
\begin{tabular}{lll}
\toprule
Feature                         & wrapper & filter \\
\midrule
taxesAndInsurance               & 0                                    & 0.095                                      \\
superConformingFlag             & 0                                    & 0.004                                      \\
repurchaseFlag                  & 0                                    & 0.006                                      \\
remainingMonthToLegalMaturity   & 0                                    & 0.033                                      \\
originalLoanTerm                & 0                                    & 0.053                                      \\
numberOfBorrowers               & 0                                    & 0.024                                      \\
monthlyReportingPeriod          & 0                                    & 0.094                                      \\
modificationCost                & 0                                    & 0.013                                      \\
miscellaneousExpenses           & 0                                    & 0.012                                      \\
miRecoveries                    & 0                                    & 0.075                                      \\
maintenanceAndPreservationCosts & 0                                    & 0.096                                      \\
loanAge                         & 0                                    & 0.092                                      \\
estimatedLoandToValue           & 0                                    & 0                                          \\
sellerName                      & 0                                    & 0.00997                                    \\
ServicerName                    & 0                                    & 0.00603     \\
\bottomrule
\end{tabular}
\end{table}

Furthermore, we labeled the customers as default vs non-default by adding a new feature, \textit{defaulted}, from the feature \textit{zeroBalanceCode} (see section \ref{subsubsection:data_labeling}). This also allows us to remove the feature \textit{zeroBalanceCode} as it is already the part of the response/target variable \textit{defaulted}.

\subsubsection{Oversampling minority class using SMOTE}\label{subsubsection:oversampling_minority_class}
For the implementation of the SMOTE oversampling technique, we installed the Python SMOTE package (i.e., imblearn). We only oversampled the minority class of the training dataset after the training/test split because oversampling the minority class of the whole dataset before splitting creates additional problems. For instance, a lot of same/similar minority examples are repeated in both the training and test set, making the model overfit the data. For this reason, the test set also becomes familiar to the model and when the model is tested against the test set, it finds a lot of common patterns. In this case, the classifier performance metrics will show overly promising performance which is actually not true as test data is not representative of real-world data, as it contains a lot more synthetic minority class. The real performance can be realized when the model is trained on oversampled data but tested on a new test set with relatively unfamiliar data (e.g., real-world skewed data) and not oversampled. Finally, after oversampling the minority class in the training set, the class ratio of default vs non-default samples in training set become equal. 

\subsubsection{Holdout set}\label{subsubsection:holdout_set}
For the purpose of testing the model's actual performance, we kept 30\% of the original data out of the training data. This separated portion of the dataset (i.e., holdout set) was not over-sampled and was not shown to the model in the training phase. We repeated this process for each of the 19 years of data. 

\subsection{Different loan underwriting regimes, stages of the economy, and associated default rate}\label{subsec:different_loan_underwriting}
\citeauthor{fout2018credit} (\citeyear{fout2018credit}), argue that there are three different loan underwriting standards and economic environments during the 19 years of data: (1) 2002 to 2004 (early housing boom), (2) 2005-2007 (late housing boom), and (3) 2011 to 2013 (post-financial crisis period). In similar fashion, by analyzing the data, we also found that we can roughly divide the loan origination vintages in to three different regimes: (1) 1999 to 2004, loan origination vintage with medium default rate (i.e., avg. 2.33\%); (2) 2005 to 2010, loan origination vintage with high default rate that includes the years of global financial crisis (i.e., avg. 4.05\%); and (3) 2011 to 2017, loan origination vintage with low default rate (i.e., avg. .21\%). As there are multiple (on average 45) performance records against one loan origination record for a customer, and a customer is identified as default in one of those performance records. After concatenating these two kinds of records, the actual ratios of the default sample (i.e., positive) reduce into much smaller ratios. Therefore, after concatenating the loan origination and performance files, the average default rate in the dataset for 1999-2004 becomes .05\%, 2005 to 2010 becomes .09\%, and for 2011 to 2017 .01\%. So the new range of defaults over all vintage becomes 0.01\% to .18\%. In any of the vintage years, the maximum positive sample (default is represented by 1 in our experiment) is .18\%, which is why we call this imbalance extreme. The main purpose of dividing all vintage years into three regimes is to see the sensitivity (i.e., changes in the recall) of different models over the difference in positive sample percentage over different default rate regimes. In short, we will call these three regimes medium, high, and low (i.e., the default rate regime) for the rest of the study.

\subsection{Evaluation criteria}\label{subsec:evaluation_criteria}
Most of the performance metrics consist of the elements from the confusion matrix. Table \ref{tab:confusion_matrix} is the confusion matrix for a binary classification problem.
\begin{table}

\centering
\caption{Confusion Matrix: columns represent ground truth, whereas the rows represent predicted result. }
\label{tab:confusion_matrix}
\begin{tabular}{lll}
\toprule
              & Actual (1)                          & Actual (0)                          \\
\midrule
Predicted (1) & True Positive (TP)                                                            & \begin{tabular}[c]{@{}l@{}}False Positive (FP) \\/ Type I Error\end{tabular}  \\
Predicted (0) & \begin{tabular}[c]{@{}l@{}}False Negative (FN) \\/ Type II Error\end{tabular} & True Negative (TN)      \\
\bottomrule
\end{tabular}
\end{table}
True Positive (TP) is the number of samples that are actually positive and predicted by the model as positive.
False Positive (FP) is the number of samples sample that is actually negative but predicted by the model as positive. It is also called Type I error.
False Negative (FN) is the number of samples that are actually positive but predicted by the model as negative. It is also called Type II error.
True Negative (TN) is the number of samples that are actually negative and predicted by the model as negative. 

\begin{itemize}
    \item Accuracy = (TP+TN) / (TP+TN+FP+FN) 
    \item Precision or Positive Predictive Value = TP / (TP + FP) 
    \item Recall or Sensitivity or True Positive Rate (TPR) = TP/ (TP+ FN) 
    \item False Positive Rate (FPR) = FP / (FP+TN)
\end{itemize}

For, imbalanced data/class, accuracy is not good to measure the performance, as it can simply classify all samples as the majority class and gain the accuracy equal to the percentage of the majority samples. As our dataset is highly imbalanced, we avoided using accuracy as a performance metric. The default account is labeled as positive (i.e., 1) in our dataset and is the target classification. So, recall/sensitivity/TPR are better metrics for the imbalanced dataset.  Precision is also important as it is a measure of the positive predictive rate. However, in the case of the imbalanced dataset, a high precision can be achieved by focusing more on classifying negative samples (e.g., non-default), where the number of FPs reduces and the ratio TP/(TP + FP) could ended up giving a higher value. Therefore, precision is less important than recall for our case. 

Another important metric is Area Under Curve (AUC), or Receiver Operating System (ROC), which is measured by taking the covered area of the ROC space by observing the TPR vs FPR for different classification thresholds. This is a standard metric used in the literature for imbalanced data (\citeauthor{garcia2012improving}, \citeyear{garcia2012improving}; \citeauthor{brown2012experimental}, \citeyear{garcia2012improving}). In our experiments, we calculate AUC from the area of ROC space.  Thus, a good performance fit is a higher TPR and lower FPR. Finally, we emphasize the performance metric \textit{recall}, AUC, and precision in order for our entire experimental result comparison.

\section{Results and discussion}\label{sec:results_and_descussions}
\begin{table*}
\caption{Ranking of algorithms for entire loan vintage periods (1999\textemdash{}2017). The appended ''-R'' with algorithm name refers to the result using resampled training data on the same holdout set.}
\label{tab:ranking_entire_vintage}
\centering
\begin{tabular}{lllllllll}
\toprule
\multicolumn{3}{l}{Rank by Precision} & \multicolumn{3}{l}{Rank by Recall} & \multicolumn{3}{l}{Rank by ROC-AUC}  \\
R\# & Alg.  & Prec.                   & R\# & Alg.  & Recall               & R\# & Alg.  & AUC                    \\
\midrule
1   & GA    & 0.947                   & 1   & SVM-R & 0.922                & 1   & GB-R  & 0.998                  \\
2   & GA-R  & 0.947                   & 2   & GA    & 0.871                & 2   & SVM-R & 0.998                  \\
3   & AB-R  & 0.836                   & 3   & GA-R  & 0.871                & 3   & SVM   & 0.996                  \\
4   & AB    & 0.8                     & 4   & ANN-R & 0.84                 & 4   & AB    & 0.988                  \\
5   & MDA-R & 0.783                   & 5   & GB-R  & 0.817                & 5   & NB    & 0.97                   \\
6   & GB-R  & 0.752                   & 6   & SVM   & 0.799                & 6   & AB-R  & 0.966                  \\
7   & ET    & 0.737                   & 7   & LR-R  & 0.796                & 7   & GA-R  & 0.957                  \\
8   & ET-R  & 0.737                   & 8   & MDA-R & 0.795                & 8   & ET-R  & 0.942                  \\
9   & DT    & 0.727                   & 9   & MDA   & 0.795                & 9   & GA    & 0.939                  \\
10  & GB    & 0.727                   & 10  & AB-R  & 0.791                & 10  & NB-R  & 0.939                  \\
11  & RF-R  & 0.702                   & 11  & DT    & 0.772                & 11  & ET    & 0.937                  \\
12  & MDA   & 0.697                   & 12  & AB    & 0.768                & 12  & LR-R  & 0.936                  \\
13  & RF    & 0.684                   & 13  & DT-R  & 0.734                & 13  & MDA-R & 0.924                  \\
14  & LR    & 0.664                   & 14  & NB-R  & 0.705                & 14  & ANN-R & 0.924                  \\
15  & DT-R  & 0.623                   & 15  & GB    & 0.687                & 15  & MDA   & 0.923                  \\
16  & ANN-R & 0.571                   & 16  & ET    & 0.658                & 16  & RF-R  & 0.91                   \\
17  & NB    & 0.526                   & 17  & ET-R  & 0.646                & 17  & RF    & 0.906                  \\
18  & NB-R  & 0.413                   & 18  & LR    & 0.637                & 18  & GB    & 0.889                  \\
19  & RS-R  & 0.364                   & 19  & RF-R  & 0.611                & 19  & DT    & 0.887                  \\
20  & SVM   & 0.345                   & 20  & ANN   & 0.558                & 20  & LR    & 0.879                  \\
21  & ANN   & 0.282                   & 21  & RS    & 0.556                & 21  & DT-R  & 0.868                  \\
22  & SVM-R & 0.246                   & 22  & RF    & 0.546                & 22  & ANN   & 0.75                   \\
23  & LR-R  & 0.192                   & 23  & RS-R  & 0.538                &     &       &                        \\
24  & RS    & 0                       & 24  & NB    & 0.445                &     &       &        \\
\bottomrule
\end{tabular}
\end{table*}

From the analysis of 19 individual mortgage loan vintage years (1997\textemdash{}2017), we found that SVM-R exhibits the highest average recall of .922 on the holdout set. In terms of precision, GA (both GA and GA-R) exhibits the highest average precision of .947 on the holdout set. Furthermore, in terms of AUC, GB-R and SVM-R were on the top of the list with a value of .998. Overall, as we are concerned about recall, followed by the AUC, and then precision, SVM-R (using resampled data) is a clear winner. On the other hand, RS, NB, and ANN are at the bottom of the list in terms of precision, recall, and AUC accordingly, as these algorithms are more sensitive to class imbalance. Interestingly, none of the algorithms which are trained on resampled data are in the bottom of the list, demonstrating that resampling improves results in terms of our desired metrics. Detail results are in Table \ref{tab:ranking_entire_vintage}, and summary results are in Table \ref{tab:best_worst_alg_entire}. RS doesn't give the confidence value (i.e., probability of being part of a particular class) of the decision. Instead, it just gives the binary decision. Thus, it was not possible to measure the AUC for RS, which is the reason for the two blank rows for the AUC column in Table \ref{tab:ranking_entire_vintage}, \ref{tab:ranking_med_vintage}, \ref{tab:ranking_high_vintage}, and \ref{tab:ranking_low_vintage}. 
\begin{table}
\caption{Best and worst algorithm with the corresponding value for metrics precision, recall, and AUC among the entire loan vintage periods (1999\textemdash{}2017).}
\label{tab:best_worst_alg_entire}
\centering
\begin{tabular}{lll}
\toprule
Metrics   & Average Value & Best Algorithm  \\
\midrule
Precision & 0.947         & GA, GA-R        \\
Recall    & 0.922         & SVM-R           \\
ROC-AUC   & 0.998         & GB-R, SVM-R    \\
\bottomrule
\end{tabular}
\end{table}

\begin{table}
\caption{The average performance of all algorithm on the same holdout set, however, trained using both original data and resampled data, for entire loan vintage years 1999\textemdash{}2017.}
\label{tab:avg_per_entire}
\centering
\begin{tabular}{lll}
\toprule
Metrics (Avg.) & Using Orig. Data & Using Res. Data  \\
\midrule
Precision      & 0.594784779              & 0.597205384                \\
Recall         & 0.674305624              & 0.755501024                \\
ROC-AUC        & 0.908279042              & 0.942057291               \\
\bottomrule
\end{tabular}
\end{table}

\begin{figure}
  \centering
  \includegraphics[width=\linewidth]{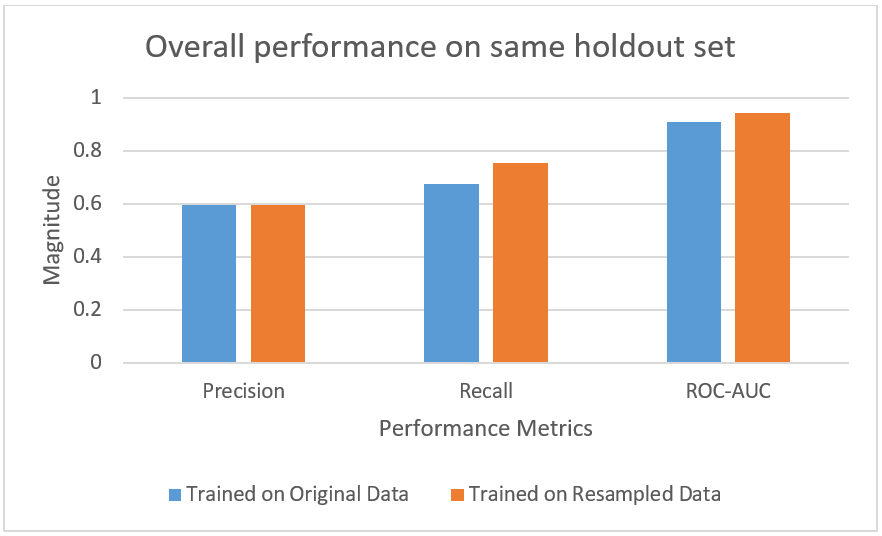}
  \caption{The average performance of all algorithm on the same holdout set, however, trained using both original data and resampled data, for entire loan vintage years 1999\textemdash{}2017.}
  \label{fig:overall_performance}
\end{figure}

We also compare the results of all the models trained on the original dataset vs all the models trained on the resampled dataset to see the magnitude of performance gain in terms of all metrics. After testing with the same holdout set, we found that the average over all the models (trained on resampled data) for precision, recall, and AUC outperformed the models trained on the original data, showing an improvement of .002 in precision, .081 in recall, and .034 in AUC (see Table \ref{tab:avg_per_entire} and Figure \ref{fig:overall_performance}). Around an 8\% increase of recall, with an increasing precision of around 3\%, has the potential to avoid roughly 8\%  of the total losses from mortgage defaults (according to \citeauthor{freddie_mac_summary_stat}, \citeyear{freddie_mac_summary_stat} the average yearly cumulative loss amount in the dataset due to mortgage defaults is 2 billion USD ). In other words, re-sampling data using SMOTE enables the algorithms to achieve, on average, 8\% more default account detection without increasing the false positives (in fact, it is decreased by 3\%). Without adding any new information, this further demonstrates the importance of re-sampling the minority class for highly imbalanced dataset.

We further investigate the variation of performance in different models with a change in class distribution (i.e., an abrupt change in default rate) due to different stages of the economy (i.e., typically, when the economy is good, the default rate is low, a medium economy comes with moderate default rate, and a bad economy comes with a high default rate). As discussed earlier in Section \ref{subsec:different_loan_underwriting}, we divided the 19 years of loan vintage into three regimes: medium default rate regime (1999-2004, average default sample .05\%); high default rate regime (2005-2010, average default sample .09\%) which includes global financial crisis periods 2007- 2009 and exhibits the highest default rate; and low default rate regime (2011-2017, average default sample .01\%). The range of default samples in these 19 years of data is highly skewed, somewhere between 0.0 and 0.2 percent. In all three regimes (low, medium, and high default rate regimes), models trained on re-sampled data showed promising performance compared to the models that were trained on the non-sampled data. In terms of recall, in the medium and high default regimes, SVM-R outperforms other models. While in the low default regimes, GB-R is slightly better than SVM-R, which indicates that GB is a little bit less sensitive to class imbalance than SVM. Detail results from all three regimes are shown in Table \ref{tab:ranking_med_vintage}, \ref{tab:ranking_high_vintage}, and \ref{tab:ranking_low_vintage} in Appendix \ref{app:app1}.

Table \ref{tab:diff_stage_best_alg} exhibits a summary of the best algorithms in different regimes. To reiterate, SVM-R and GB-R show the highest recall. In terms of AUC, AB, AB-R, and SVM-R were among the top. For our main performance metric of interest, recall, it is obvious that SVM outperforms all models on re-sampled data for the entire 19 years of time periods. However, when we divide the 19 years into three default rate regimes, SVM outperforms other models only for medium and high default rate regimes, and for the low default rate regime, GB is the winner using resampled data. This indicates that when the class imbalance is extreme, GB can be a good choice too given it is trained on re-sampled data.

\begin{table*}
\caption{A summary of best-performing algorithms in different stages of the economy or different ranges (low, medium, and high) of default rate vintage years, in terms of precision, recall, and AUC.}
\label{tab:diff_stage_best_alg}
\centering
\begin{tabular}{llll}
\toprule
Metrics   & Low (2011-2017)                        & Medium (1999-2004)             & High (2005-2010)  \\
\midrule
Precision & RF-R, RF, NB, AB-R, GA-R, ET-R, GA, ET & RF, NB, LR, ET, ET-R, GA, GA-R & GA, GA-R          \\
Recall    & GB-R                                   & SVM-R                          & SVM-R             \\
AUC       & AB                                     & AB-R                           & SVM-R    \\
\bottomrule
\end{tabular}
\end{table*}

Furthermore, Table \ref{tab:diff_stage_worst_alg} summarizes the worst performing algorithms. In terms of recall, RS is mostly affected (worst performing) in the highly (i.e., low default rate) and medium skewed dataset. In high default rate regime (comparatively less skewed data), NB performed worst. 

\begin{table*}
\caption{A summary of worst performing algorithms in different stages of the economy or different ranges (low, medium, and high) of default rate vintage years, in terms of precision, recall, and AUC.}
\label{tab:diff_stage_worst_alg}
\centering
\begin{tabular}{llll}
\toprule
Metrics   & Low (2011-2017) & Medium (1999-2004) & High (2005-2010)  \\
\midrule
Precision & RS              & RS                 & NB                \\
Recall    & RS              & RS                 & NB                \\
AUC       & MDR-R           & SVM                & ANN              \\
\bottomrule
\end{tabular}
\end{table*}

\begin{table}
\caption{The average execution time (over 19 individual years of data) by the different algorithms.}
\label{tab:execution_time}
\centering
\begin{tabular}{ll}
\toprule
Algorithm & Average Time (seconds)  \\
\midrule
RF        & 4.529                   \\
NB        & 7.724                   \\
DT        & 12.808                  \\
MDA       & 31.069                  \\
ET        & 35.748                  \\
GA        & 40.791                  \\
AB        & 131.996                 \\
RS        & 140.146                 \\
LR        & 165.555                 \\
ANN       & 207.301                 \\
GB        & 354.63                  \\
SVM       & 393.127                \\
\bottomrule
\end{tabular}
\end{table}

In term of execution time, we find that RF, NB, and DT are comparatively less time consuming, as opposed to SVM and GB (see Table \ref{tab:execution_time}). However, while SVM and GB are comparatively time consuming, it is not really a big issue as the transactions in bankruptcy prediction are usually on a monthly basis. On top of that, the mentioned computation time are for a commodity machine on around 90,000 records. In a real-world application, there are different ways to optimize this time using powerful computation resources and efficient techniques (e.g., high performance computing).   

To summarize, different algorithms show different levels of sensitivity based on the level of class imbalance present in the data. Overall, ''black box'' models (e.g., SVM, Ensemble techniques, GA) show a promising performance over the entire periods, as well as in the individual stages of the economy. 

\section{Conclusion}\label{sec:conclusions}
We investigate a large class of recent and popular bankruptcy prediction models for mortgage default prediction on a well-known public dataset from Freddie Mac. While dealing with extreme class imbalance and different stages of the economy, we discover that, in terms of the important performance metric for imbalanced data, almost all models tend to show better performance when the minority class of the training dataset is over-sampled using artificial/synthetic data. We also find that the level of imbalance has performance issues in different stages of the economy. We expect that this comprehensive study will help the practitioner with an understanding of the pros/cons of different tools with respect to class imbalance and economic stages for developing/tuning their BPM model. 

\subsection{Limitation and future works}\label{subsec:limitation_and_future_works}
Due to the choice of a variety of algorithms and added computing requirements, we resort to sampling techniques. While our sample is representative of actual data, replicating all experiments using the entire Freddie Mac dataset is worth investigating. To resample the minority class, we only use the well-known technique SMOTE. However, there are some other variations of SMOTE will be investigated. Other than SMOTE, Generative Adversarial Networks (GANs) has been adapted for generating synthetic data. Further investigation into different re-sampling techniques or synthetic data generation techniques will be a future direction of research. While ANN and SVM were referred to as the top two BPM in different studies (\citeauthor{alaka2018systematic}, \citeyear{alaka2018systematic}), we were only able to demonstrate the effectiveness of SVM. Usually, ANNs are known to need a large number of samples (i.e., data-hungry) to produce optimal performance (\citeauthor{kumar2011comparative}, \citeyear{kumar2011comparative}), therefore, experiments on the entire dataset might demonstrate the superiority of ANN. We also observed that ANN is very sensitive to the configuration of hyper-parameters (e.g., number of neurons, layers). Some class distributions did not perform as well, even after the use of GridSearchCV that works with different combinations of hyper-parameters and chooses the best one to fit the training data. A future direction of this work include ANN training with sufficient samples and finding an optimal configuration. However, a few of the top performing models such as SVM, ANN, GA, GB, ET, RF, and AB are ''black box'', lacking clear interpretations and explanations. Some researcers have argued that financial decisions (e.g., mortgage approval, bankruptcy prediction) need to be more than just a number or binary decision (\citeauthor{samek2017explainable}, \citeyear{samek2017explainable}). In addition, the European Union implemented the rule of ''right of explanation'', where a user can ask for an explanation of an algorithmic decision (\citeauthor{goodman2017european}, \citeyear{goodman2017european}). More recently, in a newly proposed bill by the U.S. government called the
''Algorithmic Accountability Act'' would require companies to assess their machine learning systems for bias and discrimination and take corrective measures (\citeauthor{algorithmic_accountability}, \citeyear{algorithmic_accountability}). Explainability of a ''black box model'' is an emerging area of research. In one of our recent works (\citeauthor{islam2019infusing}, \citeyear{islam2019infusing}), we demonstrate a technique to infuse domain knowledge in a ''black box'' model for better explainability with application to bankruptcy prediction. However, our approach suffers from different limitations (e.g., validation with multiple datasets and domains).Therefore, an emphasis on explainability and interpretability of black box models for bankruptcy prediction could be another future direction of this research.

\section*{Acknowledgment}
Thanks to Tennessee Tech Cyber-security Education, Research and
Outreach Center (CEROC) for funding this research. 

\bibliographystyle{cas-model2-names}

\bibliography{cas-refs}

\clearpage

\appendix
\section{Appendix}\label{app:app1}

\begin{table*}
\caption{Ranking of algorithms for medium default rate loan vintages or moderate economical stage (1999\textemdash{}2004) for metrics precision, recall, and AUC. The appended ''-R'' with algorithm name refers to the result using resampled training data on the same holdout set.}
\label{tab:ranking_med_vintage}
\centering
\begin{tabular}{lllllllll}
\toprule
\multicolumn{3}{l}{Rank by Precision} & \multicolumn{3}{l}{Rank by Recall} & \multicolumn{3}{l}{Rank by ROC-AUC}  \\
R\# & Alg.  & Prec.                   & R\# & Alg.  & Recall               & R\# & Alg.  & AUC                    \\
\midrule
1   & NB    & 1                       & 1   & SVM-R & 0.984                & 1   & AB-R  & 1                      \\
2   & LR    & 1                       & 2   & GB-R  & 0.974                & 2   & AB    & 1                      \\
3   & ET-R  & 1                       & 3   & SVM   & 0.971                & 3   & GB-R  & 1                      \\
4   & RF-R  & 1                       & 4   & DT    & 0.961                & 4   & LR-R  & 0.997                  \\
5   & GA-R  & 1                       & 5   & DT-R  & 0.961                & 5   & SVM   & 0.994                  \\
6   & RF    & 1                       & 6   & ANN-R & 0.958                & 6   & SVM-R & 0.994                  \\
7   & ET    & 1                       & 7   & GA-R  & 0.948                & 7   & NB    & 0.992                  \\
8   & GA    & 1                       & 8   & GA    & 0.948                & 8   & LR    & 0.992                  \\
9   & MDA-R & 1                       & 9   & MDA-R & 0.948                & 9   & ET-R  & 0.989                  \\
10  & MDA   & 0.98                    & 10  & ANN   & 0.948                & 10  & RF    & 0.987                  \\
11  & AB-R  & 0.976                   & 11  & NB    & 0.948                & 11  & ET    & 0.987                  \\
12  & AB    & 0.968                   & 12  & LR    & 0.948                & 12  & NB-R  & 0.986                  \\
13  & GB    & 0.958                   & 13  & ET-R  & 0.948                & 13  & RF-R  & 0.986                  \\
14  & DT    & 0.943                   & 14  & ET    & 0.948                & 14  & GB    & 0.985                  \\
15  & GB-R  & 0.915                   & 15  & MDA   & 0.948                & 15  & MDA   & 0.982                  \\
16  & DT-R  & 0.887                   & 16  & AB-R  & 0.948                & 16  & DT-R  & 0.979                  \\
17  & ANN-R & 0.838                   & 17  & AB    & 0.948                & 17  & ANN   & 0.978                  \\
18  & NB-R  & 0.785                   & 18  & GB    & 0.948                & 18  & DT    & 0.975                  \\
19  & ANN   & 0.527                   & 19  & NB-R  & 0.948                & 19  & GA    & 0.965                  \\
20  & SVM   & 0.437                   & 20  & LR-R  & 0.948                & 20  & ANN-R & 0.965                  \\
21  & RS-R  & 0.313                   & 21  & RF-R  & 0.924                & 21  & GA-R  & 0.965                  \\
22  & LR-R  & 0.267                   & 22  & RF    & 0.895                & 22  & MDA-R & 0.96                   \\
23  & SVM-R & 0.23                    & 23  & RS-R  & 0.702                &     &       &                        \\
24  & RS    & 0.001                   & 24  & RS    & 0.394                &     &       &     \\
\bottomrule
\end{tabular}
\end{table*}

\begin{table*}
\caption{Ranking of algorithms for high default rate loan vintages or bad economical stage (2005\textemdash{}2010) for metrics precision, recall, and AUC. The appended ''-R'' with algorithm name refers to the result using resampled training data on the same holdout set.}
\label{tab:ranking_high_vintage}
\centering
\begin{tabular}{lllllllll}
\toprule
\multicolumn{3}{l}{Rank by Precision} & \multicolumn{3}{l}{Rank by Recall} & \multicolumn{3}{l}{Rank by ROC-AUC}  \\
R\# & Alg.  & Prec.                   & R\# & Alg.  & Recall               & R\# & Alg.  & AUC                    \\
\midrule
1   & GA    & 0.875                   & 1   & SVM-R & 0.938                & 1   & SVM-R & 1                      \\
2   & GA-R  & 0.875                   & 2   & GA    & 0.875                & 2   & SVM   & 0.999                  \\
3   & AB-R  & 0.625                   & 3   & GA-R  & 0.875                & 3   & GB-R  & 0.997                  \\
4   & GB-R  & 0.594                   & 4   & ANN-R & 0.813                & 4   & AB    & 0.973                  \\
5   & AB    & 0.563                   & 5   & AB-R  & 0.688                & 5   & GA-R  & 0.969                  \\
6   & MDA-R & 0.51                    & 6   & GB-R  & 0.688                & 6   & NB    & 0.952                  \\
7   & DT-R  & 0.479                   & 7   & MDA-R & 0.688                & 7   & GA    & 0.937                  \\
8   & DT    & 0.469                   & 8   & MDA   & 0.688                & 8   & AB-R  & 0.921                  \\
9   & MDA   & 0.417                   & 9   & SVM   & 0.688                & 9   & ANN-R & 0.916                  \\
10  & GB    & 0.406                   & 10  & LR-R  & 0.688                & 10  & ET    & 0.906                  \\
11  & ET-R  & 0.375                   & 11  & AB    & 0.625                & 11  & ET-R  & 0.906                  \\
12  & ET    & 0.375                   & 12  & DT-R  & 0.625                & 12  & MDA   & 0.896                  \\
13  & RS-R  & 0.34                    & 13  & DT    & 0.625                & 13  & MDA-R & 0.896                  \\
14  & ANN-R & 0.332                   & 14  & RS    & 0.625                & 14  & NB-R  & 0.885                  \\
15  & RF-R  & 0.292                   & 15  & RS-R  & 0.5                  & 15  & LR-R  & 0.865                  \\
16  & NB-R  & 0.259                   & 16  & NB-R  & 0.5                  & 16  & RF-R  & 0.844                  \\
17  & RF    & 0.25                    & 17  & GB    & 0.438                & 17  & RF    & 0.844                  \\
18  & LR    & 0.229                   & 18  & ET-R  & 0.375                & 18  & DT    & 0.812                  \\
19  & SVM-R & 0.194                   & 19  & ET    & 0.375                & 19  & DT-R  & 0.812                  \\
20  & SVM   & 0.18                    & 20  & RF-R  & 0.313                & 20  & GB    & 0.781                  \\
21  & LR-R  & 0.171                   & 21  & LR    & 0.313                & 21  & LR    & 0.751                  \\
22  & ANN   & 0.025                   & 22  & RF    & 0.25                 & 22  & ANN   & 0.492                  \\
23  & RS    & 0                       & 23  & ANN   & 0.125                &     &       &                        \\
24  & NB    & 0                       & 24  & NB    & 0                    &     &       &     \\
\bottomrule
\end{tabular}
\end{table*}

\begin{table*}
\caption{Ranking of algorithms for low default rate loan vintages or comparatively good economical stage (2011\textemdash{}2017) for metrics precision, recall, and AUC. The appended ''-R'' with algorithm name refers to the result using resampled training data on the same holdout set.}
\label{tab:ranking_low_vintage}
\centering
\begin{tabular}{lllllllll}
\toprule
\multicolumn{3}{l}{Rank by Precision} & \multicolumn{3}{l}{Rank by Recall} & \multicolumn{3}{l}{Rank by ROC-AUC}  \\
R\# & Alg.  & Prec.                   & R\# & Alg.  & Recall               & R\# & Alg.  & AUC                    \\
\midrule
1   & NB    & 1                       & 1   & GB-R  & 0.91                 & 1   & AB    & 0.999                  \\
2   & AB-R  & 1                       & 2   & SVM-R & 0.862                & 2   & AB-R  & 0.999                  \\
3   & GA-R  & 1                       & 3   & MDA-R & 0.856                & 3   & SVM-R & 0.999                  \\
4   & ET-R  & 1                       & 4   & MDA   & 0.856                & 4   & GB-R  & 0.999                  \\
5   & GA    & 1                       & 5   & LR    & 0.856                & 5   & LR-R  & 0.985                  \\
6   & ET    & 1                       & 6   & LR-R  & 0.856                & 6   & NB    & 0.979                  \\
7   & RF-R  & 1                       & 7   & ANN   & 0.856                & 7   & NB-R  & 0.972                  \\
8   & RF    & 1                       & 8   & GB    & 0.854                & 8   & GB    & 0.972                  \\
9   & AB    & 0.981                   & 9   & DT    & 0.854                & 9   & LR    & 0.96                   \\
10  & GB    & 0.97                    & 10  & GA    & 0.854                & 10  & ET-R  & 0.957                  \\
11  & MDA-R & 0.965                   & 11  & AB    & 0.852                & 11  & RF-R  & 0.955                  \\
12  & MDA   & 0.965                   & 12  & GA-R  & 0.847                & 12  & GA-R  & 0.947                  \\
13  & LR    & 0.965                   & 13  & ET    & 0.841                & 13  & RF    & 0.941                  \\
14  & DT    & 0.906                   & 14  & AB-R  & 0.837                & 14  & ET    & 0.941                  \\
15  & GB-R  & 0.833                   & 15  & ANN-R & 0.828                & 15  & GA    & 0.934                  \\
16  & ANN-R & 0.779                   & 16  & NB-R  & 0.824                & 16  & MDA-R & 0.928                  \\
17  & DT-R  & 0.728                   & 17  & RF-R  & 0.819                & 17  & MDA   & 0.928                  \\
18  & SVM   & 0.725                   & 18  & ET-R  & 0.809                & 18  & DT    & 0.927                  \\
19  & ANN   & 0.533                   & 19  & NB    & 0.798                & 19  & ANN   & 0.918                  \\
20  & NB-R  & 0.436                   & 20  & DT-R  & 0.771                & 20  & ANN-R & 0.914                  \\
21  & SVM-R & 0.37                    & 21  & SVM   & 0.706                & 21  & DT-R  & 0.886                  \\
22  & RS-R  & 0.341                   & 22  & RF    & 0.697                & 22  & SVM   & 0.849                  \\
23  & LR-R  & 0.204                   & 23  & RS-R  & 0.508                &     &       &                        \\
24  & RS    & 0                       & 24  & RS    & 0.385                &     &       &      \\
\bottomrule
\end{tabular}
\end{table*}

\end{document}